%% file: nodalida2025.tex
\documentclass[11pt]{article}
\usepackage{nodalida2025}
\usepackage{times}
\usepackage{url}
\usepackage{float}
\usepackage{latexsym}
\usepackage{graphicx}
\usepackage{booktabs}
\usepackage{array}
\usepackage[T1]{fontenc}
\usepackage{amsmath}
\usepackage{sansmath} %
\usepackage{hyperref} 

\usepackage{courier} %
\usepackage[dvipsnames,table,xcdraw]{xcolor} %
\usepackage[framemethod=TikZ]{mdframed}
\newmdenv[
backgroundcolor=cyan!10, 
roundcorner=5pt, 
linewidth=1pt, 
align=center,
  font=\fontfamily{cmss}\selectfont %
]{custommdframed}
\newcommand{\at}{\makeatletter @\makeatother}

\aclfinalcopy %

\title{Entity Linking using LLMs for Automated \\Product Carbon Footprint Estimation}

\author{Steffen Castle \qquad 
Julian Moreno Schneider  \qquad 
Leonhard Hennig \qquad
Georg Rehm 
\\
Deutsches Forschungszentrum für Künstliche Intelligenz GmbH (DFKI) \\ Berlin, Germany \\
{\tt first.last\at dfki.de }
}
\date{} 

\begin{document}
\maketitle
\begin{abstract}
Growing concerns about climate change and sustainability are driving manufacturers to take significant steps toward reducing their carbon footprints. For these manufacturers, a first step towards this goal is to identify the environmental impact of the individual components of their products. We propose a system leveraging large language models (LLMs) to automatically map components from manufacturer Bills of Materials (BOMs) to Life Cycle Assessment (LCA) database entries by using LLMs to expand on available component information. Our approach reduces the need for manual data processing, paving the way for more accessible sustainability practices.
\end{abstract}

\section{Introduction}
Increasing awareness of climate change and sustainability has put pressure on manufacturers to reduce their carbon footprints. Regulation such as the EU Corporate Sustainability Reporting Directive (CSRD) and the European Green Deal has further emphasized the need for transparent and accurate environmental impact assessments. A fundamental step in this process is determining the environmental impact of a product, particularly the carbon emissions generated during production. LCA databases, such as ecoinvent \cite{ecoinvent}, provide detailed information for this purpose. However, linking the raw components of a product, represented in a BOM, to relevant entries in an LCA database remains a labor-intensive task requiring specialized knowledge of manufacturing materials and processes.

Recent advances in artificial intelligence, particularly large language models (LLMs), present an opportunity to automate this process. Because they are trained on vast amounts of information and are able to efficiently synthesize their knowledge as textual data, they may be capable of providing additional context in order to link components to their production processes in a LCA database. 

In this work, we investigate the use of LLMs to streamline the product carbon footprint estimation process by mapping components from BOMs directly to a LCA database. We propose a multi-step approach using a pretrained LLM to identify and summarize component information and mapping this summary to a database using semantic similarity.

\section{Related Work}

Semantic similarity techniques are well-explored in entity linking, where embedding models match textual mentions to corresponding database entries. Hou et al. introduced a method that enhances entity embeddings by incorporating fine-grained semantic information, thereby improving the learning of contextual commonality and achieving state-of-the-art performance in entity linking \cite{hou2020improving}. Similarly, Pereira and Ferreira proposed E-BELA, an approach that aligns vector representations of mentions and entities in a shared space using literal embeddings, facilitating effective linking through similarity metrics \cite{pereira2024ebela}. These methodologies are particularly effective in domains with straightforward entity disambiguation requirements, where minimal additional context is necessary. 

Existing methods for carbon footprint estimation rely heavily on manual mapping of product components to LCA databases. This is a challenging task, requiring both LCA expertise and specialist knowledge of components and materials. Several studies have explored machine learning approaches to partially automate this process. Flamingo \cite{balaji2023flamingo} uses semantic similarity to match end products to environmental impact factors such as carbon emissions in the ecoinvent database. It does not make use of the fine-grained component details present in a BOM, and additionally it uses a private dataset and supervised approach to train an auxiliary classifier. This classifier is used in conjunction with zero-shot semantic similarity matching to find database matches. Similarly, CaML \cite{balaji2023caml} maps product text descriptions to industry sector codes, which can be used to make a coarse-grained estimate of carbon production. 

\section{Background}
\subsection{LCA Database}

In order to determine the environmental impact of a component, the first step is to map it to its corresponding entry in the LCA database. LCA databases contain lists of manufacturing process names along with technical descriptions and other information. Our approach uses ecoinvent \cite{ecoinvent} as the primary LCA database. A sample of these two fields for illustrative purposes is shown in Figure \ref{entry}. For each entry, the database includes additional information describing process inputs and outputs and other life cycle information. The most task-relevant of this data is the environmental impact data including carbon emissions. Once a component has been mapped to a database entry, this information can be used in a relatively straightforward way to estimate the total carbon footprint of a product by summing the emissions for each component. Our method thus focuses on the main challenge to the carbon footprint estimation, which is linking the components from the BOM to the LCA database.

The current manual process for linking items to the LCA database involves two sequential stages. First, a non-specialist conducts a preliminary appraisal to filter straightforward matches. Using the item’s material and description, they generate a shortlist of initial candidates (typically around five entries). These potential matches are then reviewed by an LCA expert, who validates their accuracy. In cases of mismatches, the expert manually corrects the mappings in the second stage. While expert oversight remains essential for quality assurance, our objective is to optimize this workflow by automating the preliminary non-specialist tasks. This eliminates manual effort while preserving the critical role of expert verification.

\input{entry}

\subsection{Bills of Materials}

A BOM is an industry-standard document produced by a manufacturer listing the components that make up a product. This document often contains information such as the name of the component, the supplier of the component and the material name, when available and is often required to be produced for regulatory reasons. An example is shown in Figure \ref{bom}. Although knowledge of the main material of a component is often enough to determine a suitable match to a process in the database, the material name provided by the supplier is often an internal name for the material, a specification code, or other non-straightforward description of the material.  Due to the complexity of correctly mapping component materials to the process used to produce them, specialist knowledge is usually required.

\input{bom}

\subsection{Component Datasheets}

For some components, a technical datasheet is produced by the manufacturer. Typically, this lists properties of the component or the materials that make up the component. This information can provide useful context that helps to identify the process name for the component. Datasheets are not usually available for all components; they are only supplied by the manufacturer in some instances. We investigate including the information from datasheets in the entity mapping process.

\section{Methodology}

\begin{figure*}[h]
\centerline{\includegraphics[width=\textwidth]{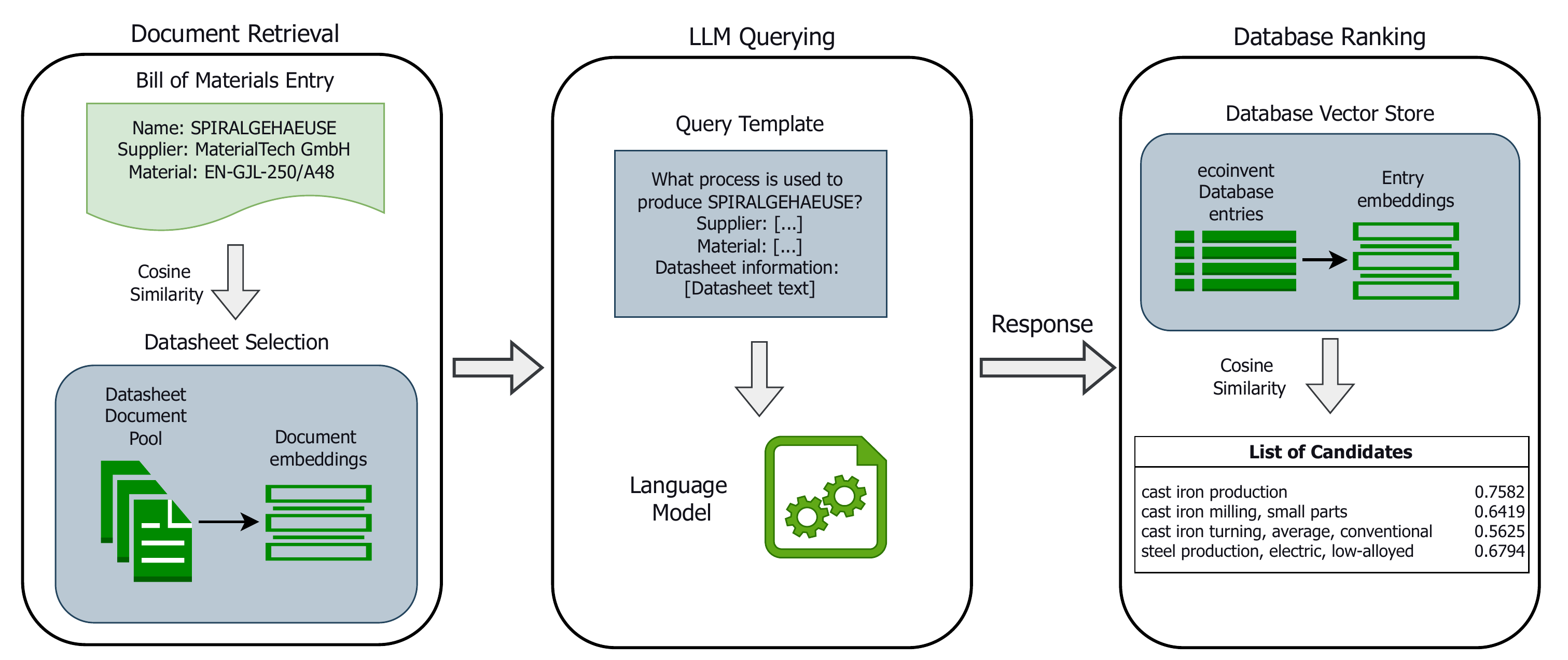}}
\caption{Architecture of the proposed pipeline, made up of three modules: Document retrieval, LLM querying, and database ranking.}
\label{pipeline}
\end{figure*}

Our key contributions are:
\begin{enumerate}
    \item Utilize fine-grained information from BOMs to provide a more accurate assessment of carbon emissions
    \item Introduce LLMs into the entity mapping process in order to provide additional context, and
    \item Integrate additional context from component datasheets in order to further improve context.
\end{enumerate}
We break the process of mapping components to database entries into three connected steps, outlined below. An illustration of the complete pipeline is shown in Figure \ref{pipeline}.

\subsection{Datasheet Selection}

Given a pool of datasheets, we must select the document from the pool corresponding to the BOM entry of interest to provide additional context about the component. In order to determine the matching document, we evaluate the textual similarity between the concatenation of the filename and text of the datasheet, and the concatenation of the component name, manufacturer and material from the BOM entry. Similarity is measured by the cosine similarity of the two embeddings generated by a text embedding model. After manual evaluation, we chose a threshold value of 0.5 or higher for the cosine similarity to indicate a match. If a match is found, the text from the datasheet is included in further processing in order to match the BOM entry to the process name.

\subsection{LLM Querying}

We utilize a LLM agent fine-tuned for chat as the LLM in our pipeline. We create a prompt for the model that includes all relevant context and instructs the model to produce a description of the manufacturing process used to create the component. Context includes the BOM entry information (component name, supplier, and material) along with the content of the datasheet, if available. The exact prompt can be found in our publicly available code\footnote{https://github.com/DFKI-NLP/eco-link}. The output of the model is then used in further processing. A sample showing typical responses from the LLM is shown in Figure \ref{response}. We use a local instance of Llama 3.1 with 8 billion parameters \cite{dubey2024llama} as the LLM in our experiments.

\input{response}

\subsection{Semantic Similarity Matching}

As a preprocessing step, we create embeddings for each database entry using the process name and description. We store these embeddings in a FAISS \cite{douze2024faiss} vector store. To find a match for a BOM entry, we create an embedding of the LLM response from the previous step and compare this embedding to the vector store. Using the cosine similarity as a distance measure, we are able to obtain a ranking of database entries. For both the datasheet selection and semantic similarity ranking, we use \texttt{gte-large-en-v1.5} \cite{li2023towards}.

\section{Results and Discussion}

\input{results}

Because of limitations in data availability due to lack of large-scale public datasets, data privacy, and preservation of trade secrecy, we can only evaluate on a small set of BOM entries. Our set of labeled evaluation data consists of only 21 components from 3 different BOMs. 

We evaluate by comparing human performance to both the full and ablated pipeline. \textbf{Semantic Similarity} uses only the Database Ranking module and semantic similarity between the database entries and the component name, supplier and material. \textbf{LLM} uses both the LLM and Database Ranking modules to match the LLM's response , and \textbf{LLM + Datasheet} includes the Document Retrieval component. We use $Hits@n$ as the metric, which is defined for a recommender system as the proportion of instances the correct item is present in the top $n$ recommendations. This metric corresponds to our use case, where a shortlist of recommendations is provided by the non-expert recommender. The results are shown in Figure \ref{results}. 

Our approach was found to be acceptable given the challenging nature of the task --- on par or slightly better than non-expert human performance. While not enough to completely automate the entire entity mapping process, these results indicate that our method could take the place of the non-expert human in this process.

\section{Conclusion and Future Work}

This paper presents a novel approach to estimate product carbon footprints using LLMs to map BOM components to LCA database entries. The proposed pipeline streamlines the traditionally manual process, achieving reasonable accuracy and scalability. Future work includes an expanded evaluation with a larger evaluation dataset and integration of additional context from sources such as web search results.
\section*{Acknowledgments}

This paper has received funding from the European Union under the grant agreement no. 101058573 (SciLake) and by the German  Bundesministerium für Umwelt, Naturschutz, nukleare Sicherheit und Verbraucherschutz (BMUV) under the Green AI Hub Mittelstand initiative.

\bibliographystyle{acl_natbib}
\bibliography{nodalida2025}

\end{document}

%% file: entry.tex
\begin{figure}[h!]

\sffamily

\begin{custommdframed}

\textbf{Steel production, electric arc furnace, EU} 

\vspace{0.3em}
This process models the production of steel using an electric arc furnace (EAF) within the European Union. The process includes the melting of recycled steel scrap and the subsequent refinement to meet industry-grade specifications. Electricity consumption and emissions are based on averages from EU-wide data. Additional inputs include limestone for slag formation and oxygen for decarburization. Outputs include steel billets ready for further processing and slag as a by-product for use in construction applications.

This dataset represents a cradle-to-gate assessment, capturing the production of steel billets up to the point of factory gate, excluding downstream processing (e.g., rolling or shaping). Energy mix and emission profiles align with EU 27 averages for 2023.
\end{custommdframed}
\rmfamily
\caption{An illustrative example of the process name and description from a LCA database.}
\label{entry}

\end{figure}

%% file: bom.tex
\begin{figure*}[h!]

\begin{custommdframed}[userdefinedwidth=38em]

\begin{tabular}{l l l}

\textbf{Component Name} & \textbf{Material} & \textbf{Supplier} \\
\hline
SPIRALGEHÄUSE & EN-GJL-250/A48 CL 35B & Mechatronik GmbH \\
WELLE & C45+N & Technikbau AG \\
SPALTRING               & JL/GUSSEISEN LAMELLENGRAFIT & GussForm Solutions   \\
SPANNRING               & STAHL+KATAPHORESE           & StahlPro Engineering \\ 
SPALTRING               & JL/GUSSEISEN LAMELLENGRAFIT & GussTech Industries  \\ 
STIFTSCHRAUBE           & 8.8                        & FixFast Components   \\ 
STIFTSCHRAUBE           & 8.8                        & SchraubenWerk AG     \\ 
STIFTSCHRAUBE           & 5.8+A2A                    & PrecisionParts GmbH  \\ 

\end{tabular}

\end{custommdframed}
\rmfamily
\caption{An excerpt from  a BOM. Note that material codes are often ambiguous or obscure, requiring specialized knowledge to correctly identify.}
\label{bom}
\end{figure*}

%% file: response.tex
\begin{figure*}[h!]
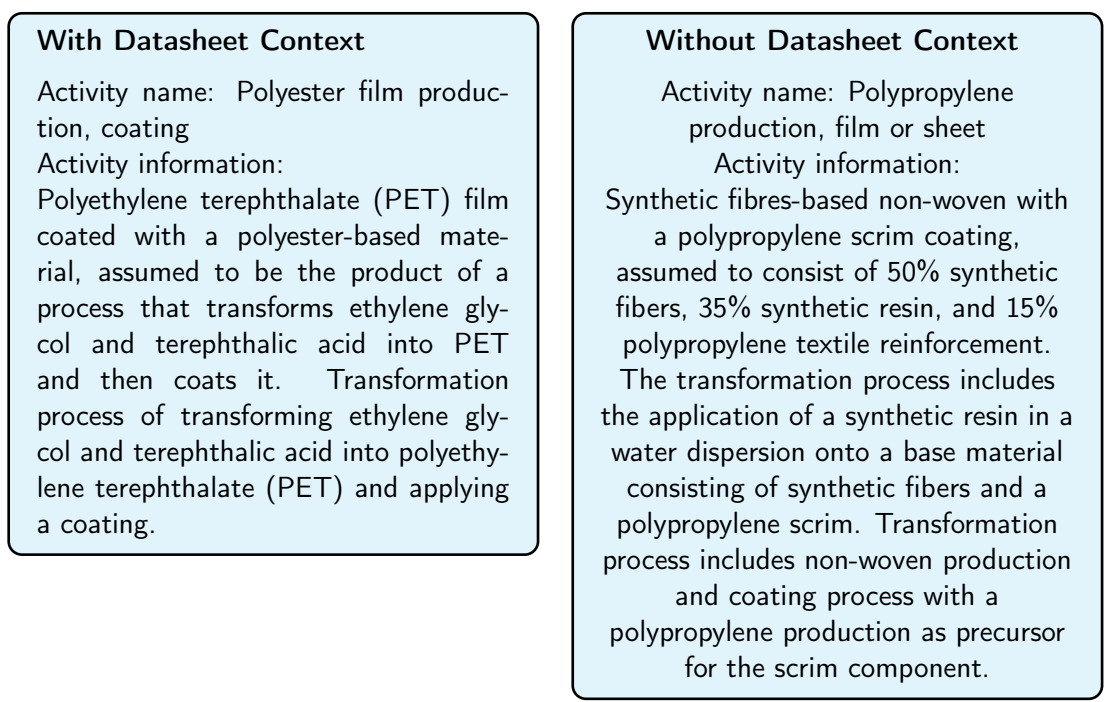

\centering
{\sffamily
\textbf{Component:} Ibitech 57 coated Normalausführung 2 mm weiss\\
(Woven polypropylene)
}
\begin{tabular}{p{7cm} p{7cm}}
\begin{custommdframed}

\textbf{With Datasheet Context}
\vspace{0.5em}

Activity name: Polyester film production, coating 

Activity information: 

Polyethylene terephthalate (PET) film coated with a polyester-based material, assumed to be the product of a process that transforms ethylene glycol and terephthalic acid into PET and then coats it. Transformation process of transforming ethylene glycol and terephthalic acid into polyethylene terephthalate (PET) and applying a coating.
\end{custommdframed}
 & 
\begin{custommdframed}
\centering
\textbf{Without Datasheet Context}
\vspace{0.5em}

 Activity name: Polypropylene production, film or sheet

Activity information: 

Synthetic fibres-based non-woven with a polypropylene scrim coating, assumed to consist of 50\% synthetic fibers, 35\% synthetic resin, and 15\% polypropylene textile reinforcement. The transformation process includes the application of a synthetic resin in a water dispersion onto a base material consisting of synthetic fibers and a polypropylene scrim.
Transformation process includes non-woven production and coating process with a polypropylene production as precursor for the scrim component.
\end{custommdframed}
\end{tabular}
\label{response}
\caption{Sample LLM response for a BOM component. Inclusion of datasheet context appears to considerably improve reliability of responses.}
\end{figure*}

%% file: results.tex
\begin{figure*}[h!]
\centering
\newcolumntype{R}[1]{>{\raggedleft\let\newline\\\arraybackslash\hspace{0pt}}m{#1}}
\begin{tabular}{|c| R{3cm} | R{3cm}|}
    \hline
    \textbf{Method} & \textbf{Hits@5} & \textbf{Hits@1} \\
     \hline
    Human (non-expert) & \textbf{0.48} & 0.19 \\
    Semantic similarity only & 0.05 & 0.00 \\
    LLM & 0.43 & 0.19 \\
    LLM + Datasheet & \textbf{0.48} & \textbf{0.24} \\
    \hline
\end{tabular}
\caption{Results of evaluation. Our proposed approach is able to match the performance of a human on the top 5 matches, and exceeds human performance on the top match.}
\label{results}
\end{figure*}